\documentclass{article}
\usepackage[preprint]{corl_2026}
\usepackage{amsmath,amssymb}
\usepackage[dvipsnames]{xcolor}
\definecolor{mydarkblue}{rgb}{0,0.08,0.45}
\usepackage{booktabs}
\usepackage{multirow,colortbl,graphicx,makecell}
\graphicspath{{figures/}}
\usepackage{arydshln}
\usepackage[capitalize]{cleveref}
\usepackage{afterpage}
\title{MotionMEM: Self-Supervised Motion Tokens as Memory for Vision-Language-Action Policies}

\author{
Hongze Fu$^{*}$ \enspace Shaolong Li$^{*}$ \enspace Byeongju Woo \enspace Yunhao Luo\\[1pt]
  \textbf{Yinpei Dai \enspace Joyce Chai$^{\dagger}$ \enspace Nima Fazeli$^{\dagger}$}\\[2pt]
  {\normalfont University of Michigan \qquad {\small $^{*}$Equal contribution \quad $^{\dagger}$Equal advising}}
}

\hypersetup{pdftitle={MotionMEM: Self-Supervised Motion Tokens as Memory for Vision-Language-Action Policies},
  pdfauthor={Hongze Fu, Shaolong Li, Byeongju Woo, Yunhao Luo,
    Yinpei Dai, Joyce Chai, Nima Fazeli}}

\begin{document}
\maketitle
\suppressfloats[t]
\begin{abstract}
Robot manipulation often requires memory of earlier events, such as how many objects have been placed into a bin.
A common memory for vision-language-action policies is a sparse set of sampled past frames.
However, frames miss what happens between the sampled instants, and denser sampling raises the computational cost with every added frame.
We introduce MotionMEM, which augments the frame memory with self-supervised motion tokens that record the changes between the sampled frames.
A motion encoder, built on a frozen video VAE, summarizes each completed window of video into a single token.
It is trained to reconstruct the rest of a window from its first frame, so the token captures what changed rather than appearance.
On four motion-centric tasks derived from RoboMME, with the memory injected by modulation, adding frames to $\pi_{0.5}$ stops helping at 45\% success, even with eight times the memory.
In contrast, MotionMEM reaches 54\% with about one sixth of those tokens.
Frozen video features and an encoder trained by the action loss alone do not close this gap.
The gain carries over to context injection on the same tasks, and to the 16 general memory tasks of RoboMME.
Project website: \url{https://hongzefu.github.io/motionmem}
\end{abstract}

\section{Introduction}
\label{sec:intro}

\begin{figure}[t]
  \centering
\begin{minipage}[t]{0.37\textwidth}\centering
{\sffamily\bfseries\scriptsize (a) Sparse frames vs.\ motion tokens}\\[3pt]
    \includegraphics[width=\linewidth]{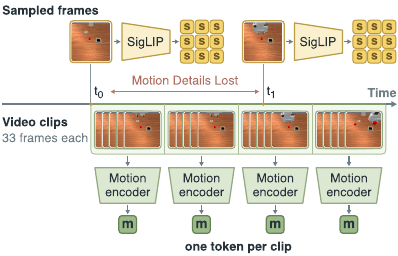}
  \end{minipage}\hspace{0.02\textwidth}\begin{minipage}[t]{0.29\textwidth}\centering
    {\sffamily\bfseries\scriptsize (b) Success vs.\ memory tokens}\\[3pt]
    \includegraphics[width=\linewidth]{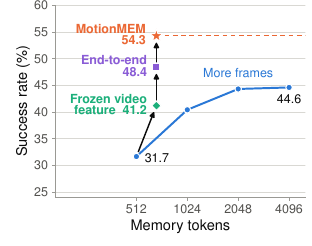}
  \end{minipage}\hspace{0.02\textwidth}\begin{minipage}[t]{0.29\textwidth}\centering
    {\sffamily\bfseries\scriptsize (c) Success by difficulty tier}\\[3pt]
    \includegraphics[width=\linewidth]{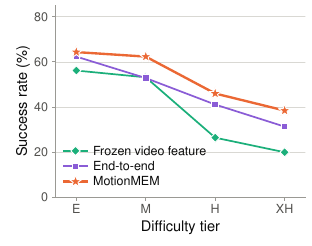}
  \end{minipage}
\caption{\textbf{Sparse frames miss what happens between them; one motion token per window records it.} \textbf{(a)} Sampled frames keep appearance at isolated instants, while a motion encoder summarizes each window (clip) into one token. \textbf{(b)} Success on the four motion-centric tasks against the number of memory tokens (modulation injection; mean of three evaluation seeds). More frames saturate. At the same budget, a frozen video feature (Wan-1, our baseline that pools the latents of the Wan video VAE~\cite{wan2025} into one token per window), an encoder trained end to end (E2E-Enc), and MotionMEM improve in turn. The dashed line marks MotionMEM. \textbf{(c)} The same three memories by difficulty tier, pooled over the four tasks. MotionMEM's margin over the frozen feature is largest on the hard and extra-hard tiers.}
  \label{fig:teaser}\label{fig:budget}
\end{figure}

Many manipulation tasks cannot be solved from the current observation alone. A robot asked to retrace a route it was shown, or to find an object after two objects have swapped places must remember what happened earlier. Vision-language-action (VLA) models, however, act mostly on the current or very recent observations, which has motivated explicit memory for robot control~\cite{dai2026robomme}. A widely used memory mechanism is a sparse set of sampled frames~\cite{dai2026robomme}, which preserve appearance, object identity, and spatial layout. However, frames are snapshots, so when two frames look alike, the policy cannot tell which process connected them (\cref{fig:teaser}a) -- the policy needs to remember the process, not just snapshots.

A straightforward fix is to sample more frames, but every added frame adds computation. On our motion-centric tasks, the gain from adding frames levels off (\cref{fig:teaser}b). Another fix is to add the frozen latents of a pretrained video VAE, one token per window, but on our motion-centric tasks neither is enough. Our key idea is instead to learn a memory of change: a token that describes how a short window of video differs from its first frame. To this end, we introduce \textit{MotionMEM}, which augments the sparse frame memory of the base policy with self-supervised motion tokens (\cref{fig:overview}). A motion encoder, built on a frozen video VAE, compresses each window into a single motion token. It is trained by reconstruction, where a decoder must recover the rest of the window from the token given the first frame as an anchor. Since the anchor frame supplies visual appearance, the token is pushed to encode what changes across frames.

We evaluate our method with $\pi_{0.5}$ on four motion-centric tasks that we derive from RoboMME~\cite{dai2026robomme} and on the 16 general memory tasks of RoboMME. In summary, we propose MotionMEM, a self-supervised motion memory for VLA policies, and show that:
\begin{itemize}\setlength{\itemsep}{1pt}\setlength{\parskip}{0pt}\setlength{\topsep}{2pt}
  \item \textbf{One learned token per window serves the policy better than more frames.} When the deciding evidence is a process and the memory is injected by modulation, an eightfold larger frame memory raises success only to 44.6\% (\cref{fig:teaser}b), whereas MotionMEM reaches 54.3\% with about one sixth of those tokens.
  \item \textbf{How the token is learned matters.} At the same budget, frozen video features and the same encoder trained by the action loss alone both fall short of the token learned by anchor-conditioned reconstruction.
  \item \textbf{The gain transfers, and the policy acts on the token.} It also appears under context injection and on the 16 general memory tasks of RoboMME, and replacing the motion tokens at inference changes which sub-goal the robot executes and when it acts (\cref{sec:interp}).
\end{itemize}
 \section{Related Work}
\label{sec:related}

\noindent\textbf{Memory for Robotic Policies.}
Memory lets a policy act on information absent from the current observation, such as earlier object states and completed interactions. Early designs extend the observation window~\cite{torne2025longcontext,jang2025contextvla,li2025cronusvla}, add a recurrent module state~\cite{li2023roboflamingo}, maintain a neural map~\cite{dai2024think}, or keep a memory bank~\cite{shi2026memoryvla,fang2025sam2act}. RoboMME~\cite{dai2026robomme} compares symbolic, perceptual, and recurrent memory on one backbone on memory-dependent manipulation. Recent work makes the history compact. LTM~\cite{sun2026ltm} encodes adjacent observation pairs with a frozen latent-action encoder~\cite{gao2026dreamdojo}, and IMPRINT~\cite{wu2026imprint} fuses multimodal histories through a pretrained recurrent bottleneck. T$^2$Mem~\cite{liu2026t2mem} writes history into fast weights through self-supervised test-time updates, and RB-VLA~\cite{bagaria2026rbvla} carries a recursive belief. HAMLET~\cite{koo2025hamlet} reduces each past step to a few learned tokens, and NativeMEM~\cite{wang2026nativemem} each past frame to a single token. Others keep the visual history itself: Divide-and-Remember~\cite{yu2026divideandremember} selects an action-relevant subset of visual tokens, and SimpleMemVLA~\cite{yin2026simplememvla} feeds timestamped frames to the backbone as native video. MotionMEM also keeps sampled frames as the appearance memory and adds a window-level motion encoder beside them. LTM is the closest, as it also records change, but with a frozen encoder pretrained on external video; our encoder is trained on the task data, on a frozen video VAE, and we compare with the number LTM reports (\cref{sec:conclusion}).

\noindent\textbf{Language-Based and Hierarchical Memory.}
A second approach keeps history as language or in a high-level module~\cite{dai2025racer,shi2025hi,hu2026matters}. MEM~\cite{physicalintelligence2026mem} maintains a compressed language record, PAL~\cite{liu2026pal} an event list and a subtask plan, and LogLook~\cite{jiang2026loglook} an execution log that a planner re-reads. MemER~\cite{sridhar2025memer} lets a high-level VLM select keyframes and issue instructions, and PonderPounce~\cite{choi2026ponderpounce} passes cognition features from a separate context model. All interpret history upstream of low-level control. MotionMEM augments the visual history with motion tokens and adds no language summary or planner.

\noindent\textbf{Self-Supervised Representations.}
VLA policies typically read images through encoders pretrained with language supervision, such as CLIP~\cite{radford2021clip} and SigLIP~\cite{zhai2023siglip}. Joint-embedding predictive architectures instead predict missing content in representation space, on images~\cite{assran2023ijepa} and on video~\cite{assran2025vjepa2}, and LeWorldModel~\cite{maes2026leworldmodel} trains such a model end to end from pixels with a Gaussian regularizer. Our motion encoder is also trained without labels, with a regularizer of this kind, SIGReg~\cite{balestriero2025lejepa}, but it reconstructs the later latents of a window given the first frame, so a single token keeps the change. Latent-action models also learn such tokens from video without action labels, but as targets for policy pretraining~\cite{ye2025lapa,chen2024moto,bu2025univla}; ours serves as policy memory.

\noindent\textbf{Interpretability of VLA Policies.}
Interpretability analyses of VLA policies measure how actions depend on visual tokens~\cite{luo2026lookbeforeacting,zhang2026focusvla}, trace representations across layers and checkpoints~\cite{shi2026vlatrace}, and attribute actions to image regions~\cite{zhang2026embodiedinterp}, but none examines what a policy retains from earlier steps. Memory papers inspect it mostly through attention over stored history~\cite{koo2025hamlet,wang2026nativemem}, with isolated probes~\cite{wu2026imprint} or content swaps~\cite{liu2026t2mem}. We apply attention readouts, token replacement, and linear probes to a learned memory (\cref{sec:interp}).
 \section{MotionMEM}
\label{sec:method}

\begin{figure*}[t]
  \centering
  \includegraphics[width=\linewidth]{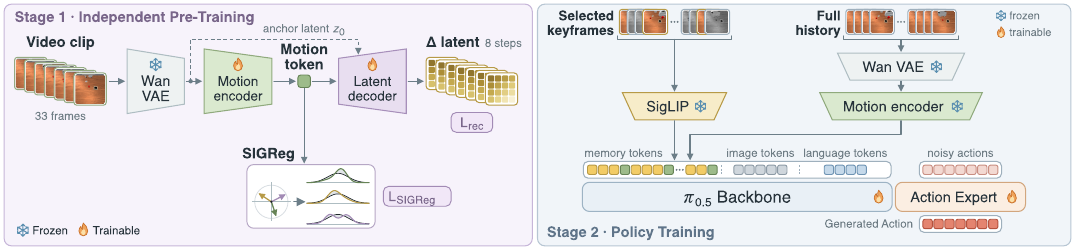}
  \caption{\textbf{Overview of MotionMEM.} Stage 1 trains the motion encoder on frozen video-VAE latents with latent reconstruction and SIGReg regularization. Stage 2 freezes the encoder and conditions $\pi_{0.5}$ on 512 SigLIP frame tokens plus at most 160 motion tokens. The panel shows context injection; with modulation the memory conditions the action expert instead.}
  \label{fig:overview}
\end{figure*}

We consider a VLA policy that predicts an action chunk $a_{t:t+H}\sim\pi(\cdot\mid \ell, o_t, \mathcal{M}_t)$ from the instruction $\ell$, the current observation $o_t$, and a memory $\mathcal{M}_t$ of the episode so far. Our goal is a memory that records both what past scenes looked like and how they changed, within a fixed token budget. MotionMEM builds $\mathcal{M}_t$ from two streams (\cref{fig:overview}): appearance tokens from sparsely sampled frames, and motion tokens from a self-supervised \emph{motion encoder} (\cref{sec:method_encoder}). The \emph{VLA policy} reads both streams as one time-ordered sequence (\cref{sec:method_policy}). We instantiate the policy on $\pi_{0.5}$~\cite{black2025pi05}.

\subsection{Motion Encoder}
\label{sec:method_encoder}

\noindent\textbf{Architecture.}
The encoder should summarize a short window of video in one token, without storing its appearance again. To this end, we build it on the frozen Wan video VAE~\cite{wan2025}. Given a window of $T$ frames, the VAE produces latents $z_{0:K}$, where $z_0$ encodes the first frame of the window, which we call the anchor. We project each latent to one token and process the tokens with a causal Transformer, so the token of $z_k$ attends only to $z_{\le k}$. The output at the last position has attended to the whole window, and we project it to a single motion token $m = \mathrm{enc}_\theta(z_{0:K})\in\mathbb{R}^{d}$. A decoder, used only during training, reconstructs the later latents from the anchor and the token, each as a residual on the anchor latent:
\begin{equation}
  \hat z_k = z_0 + \mathrm{dec}_\phi(z_0, m, k),\qquad k=1,\dots,K.
\end{equation}

\noindent\textbf{Training objective.}
We train the encoder and decoder end to end with a reconstruction loss and an anti-collapse regularizer:
\begin{equation}
  \mathcal{L}_{\mathrm{enc}}
  = \mathcal{L}_{\mathrm{rec}}
  + \lambda\,\mathrm{SIGReg}(\{m_i\}),
\end{equation}
where $\mathcal{L}_{\mathrm{rec}}$ is the mean squared error between $\hat z_k$ and $z_k$ over the later latents. Intuitively, the anchor already supplies appearance, so the only way to lower $\mathcal{L}_{\mathrm{rec}}$ is for $m$ to encode what changes within the window. SIGReg~\cite{balestriero2025lejepa} drives the batch of tokens $\{m_i\}$ toward an isotropic Gaussian and prevents collapse. The objective uses no action or language supervision. After training, we freeze $\mathrm{enc}_\theta$ and discard the decoder. We use $T=33$, $K=8$, $d=768$, and $\lambda=10^{-4}$.

\subsection{VLA Policy}
\label{sec:method_policy}

\noindent\textbf{Memory assembly.}
$\mathcal{M}_t$ is a sequence of memory tokens, each indexed by the episode timestep it describes. For appearance, we follow RoboMME's frame sampling~\cite{dai2026robomme}: $N$ frames are sampled evenly from the history, $\mathcal{S}_t=\mathrm{EvenSample}(\{0,\dots,t\},N)$, and $v_\tau$ denotes the visual tokens of frame $\tau\in\mathcal{S}_t$. For motion, we encode every completed window, taken at a fixed stride within each segment of an episode (the demonstration, if any, and the execution). Let $\mathcal{W}_t$ be the set of window starts $s$ on this stride with $s+T-1\le t$; $\mathrm{enc}_\theta$ maps the window starting at $s\in\mathcal{W}_t$ to the motion token $m_s$. We order both streams by timestep and merge them into a fixed budget of $B=B_{\mathrm{img}}+B_{\mathrm{mot}}$ slots:
\begin{equation}
\resizebox{0.9\linewidth}{!}{$\displaystyle\mathcal{M}_t=\mathrm{sort}_{\mathrm{time}}\big([\,v_{\tau_1};\dots;v_{\tau_N};\,m_{s_1};\dots;m_{s_J}\,]\big),\quad \tau_i\in\mathcal{S}_t,\ s_i\in\mathcal{W}_t.$}
\end{equation}
Each motion token is placed at the start time of the window it summarizes. We use $B_{\mathrm{img}}=512$ and $B_{\mathrm{mot}}=160$ throughout, with a stride of 16 frames; the frame count and the tokens per frame are given in \cref{sec:setup}.

\noindent\textbf{Memory injection.}
We inject $\mathcal{M}_t$ in one of the two ways introduced by RoboMME~\cite{dai2026robomme}. In the \emph{context} variant, we project each memory token to the backbone width and place the sequence, in timestep order, at the front of the prefix, before the current-observation and instruction tokens. The instruction tokens and the action expert then read the memory through the backbone's ordinary attention, which adds no parameters beyond the memory projections. In the \emph{modulation} variant, the memory adds no context tokens and instead conditions the action expert through adaptive LayerNorm. We evaluate both variants on the motion-centric tasks and only context on RoboMME (\cref{sec:setup}). We fine-tune the policy with the flow-matching action loss, keeping the image tower and $\mathrm{enc}_\theta$ frozen and updating the language backbone, action expert, and memory projections.
 \section{Experimental Setup}
\label{sec:exp}\label{sec:setup}

The experiments answer three questions: (i) when the deciding evidence is a process, does one learned token per window serve the policy better than more frames; (ii) must the token be learned with a motion-specific objective, or do frozen video features and the action loss alone suffice; and (iii) does the gain hold across injection mechanisms and on general memory tasks? \cref{sec:rq1} takes them in turn, and \cref{sec:interp} asks whether the policy uses the tokens.

\begin{table*}[t]
  \centering
\caption{\textbf{Success rates (\%, $\uparrow$) on the four motion-centric tasks.} E, M, H, and XH are difficulty tiers; Avg pools a task and All pools the 700 episodes. $B$ is the history token budget. The upper and lower blocks inject the memory by modulation and as context. Cells are the mean over three evaluation seeds, and \cref{tab:longcontext_std} adds the standard deviations. \textbf{Bold}: highest per column.}
  \label{tab:longcontext}
\setlength{\tabcolsep}{3.0pt}\renewcommand{\arraystretch}{1.05}\setlength{\aboverulesep}{0.2ex}\setlength{\belowrulesep}{0.3ex}
  \resizebox{\textwidth}{!}{\begin{tabular}{@{}lcccccccccccccccccccc@{}}
\toprule
  &   & \multicolumn{4}{c}{\cellcolor{gray!15}BinFill} & \multicolumn{5}{c}{\cellcolor{gray!15}RouteStick} & \multicolumn{4}{c}{\cellcolor{gray!15}VideoRepick} & \multicolumn{5}{c}{\cellcolor{gray!15}VideoUnmaskSwap} &   \\
\cmidrule(lr){3-6}\cmidrule(lr){7-11}\cmidrule(lr){12-15}\cmidrule(lr){16-20}
Method & $B$ & E & M & H & Avg & E & M & H & XH & Avg & E & M & XH & Avg & E & M & H & XH & Avg & All \\
\midrule
\multicolumn{21}{@{}l}{\textbf{Modulation injection}} \\
\multicolumn{21}{@{}l}{\textit{Frames}} \\
\quad FrameSamp & 512 & 80.7 & 57.3 & 16.0 & 51.3 & 84.0 & 44.7 & 17.3 & 0.7 & 36.7 & 24.0 & 28.0 & 22.7 & 24.9 & 18.0 & 26.0 & 6.7 & \textbf{17.3} & 17.0 & 31.7 \\
\quad FrameSamp & 1024 & 78.0 & 52.7 & 23.3 & 51.3 & 96.7 & 88.0 & 58.0 & 5.3 & 62.0 & 27.3 & 32.7 & 31.3 & 30.4 & 23.3 & 29.3 & 7.3 & 12.7 & \textbf{18.2} & 40.4 \\
\quad FrameSamp & 2048 & 80.7 & 68.0 & 22.7 & 57.1 & \textbf{100.0} & 95.3 & 70.0 & 38.0 & 75.8 & 30.0 & 30.7 & 24.0 & 28.2 & 16.0 & 24.7 & 6.7 & 14.0 & 15.3 & 44.3 \\
\quad FrameSamp & 4096 & 86.7 & 62.0 & 30.0 & 59.6 & 98.7 & \textbf{99.3} & 82.7 & 28.0 & 77.2 & 19.3 & 26.7 & 31.3 & 25.8 & 24.0 & 22.7 & 4.0 & 9.3 & 15.0 & 44.6 \\
\addlinespace[0.25ex]
\hdashline[0.6pt/1.6pt]
\addlinespace[0.25ex]
\multicolumn{21}{@{}l}{\textit{Frozen latents}} \\
\quad Wan-1 & 512+160 & 85.3 & 62.7 & 20.7 & 56.2 & 94.0 & 94.0 & 51.3 & 20.0 & 64.8 & 22.7 & 30.0 & 26.0 & 26.2 & 22.7 & 26.0 & 7.3 & 14.0 & 17.5 & 41.2 \\
\quad Wan-9 & 512+720 & 82.7 & 52.7 & 18.0 & 51.1 & 95.3 & 87.3 & 44.7 & 3.3 & 57.7 & 24.0 & 24.7 & 24.7 & 24.4 & 22.0 & 24.7 & 10.0 & 12.0 & 17.2 & 37.6 \\
\addlinespace[0.25ex]
\hdashline[0.6pt/1.6pt]
\addlinespace[0.25ex]
\multicolumn{21}{@{}l}{\textit{End-to-end}} \\
\quad E2E-Enc & 512+160 & 88.7 & 72.7 & 32.0 & 64.4 & 98.7 & 96.7 & \textbf{85.3} & 46.0 & 81.7 & 30.0 & 16.7 & 40.0 & 28.9 & \textbf{32.0} & 25.3 & 6.0 & 8.0 & 17.8 & 48.4 \\
\addlinespace[0.25ex]
\hdashline[0.6pt/1.6pt]
\addlinespace[0.25ex]
\multicolumn{21}{@{}l}{\textit{Reconstruction (ours)}} \\
\quad MotionMEM & 512+160 & \textbf{92.7} & \textbf{86.7} & \textbf{38.7} & \textbf{72.7} & \textbf{100.0} & 94.7 & 84.7 & \textbf{56.0} & \textbf{83.8} & \textbf{48.7} & \textbf{36.7} & \textbf{50.0} & \textbf{45.1} & 16.0 & \textbf{31.3} & \textbf{14.7} & 9.3 & 17.8 & \textbf{54.3} \\
\midrule
\multicolumn{21}{@{}l}{\textbf{Context injection}} \\
\multicolumn{21}{@{}l}{\textit{Frames}} \\
\quad FrameSamp & 512 & 76.0 & 25.3 & 8.0 & 36.4 & 28.7 & 16.7 & 7.3 & 0.0 & 13.2 & 33.3 & 24.0 & 24.0 & 27.1 & 24.7 & 20.0 & 10.0 & 10.0 & 16.2 & 22.0 \\
\quad FrameSamp & 1024 & 59.3 & 13.3 & 4.0 & 25.6 & 62.0 & 42.0 & 9.3 & 0.0 & 28.3 & 22.0 & 11.3 & 16.0 & 16.4 & 20.0 & 22.7 & 7.3 & \textbf{17.3} & 16.8 & 21.9 \\
\addlinespace[0.25ex]
\hdashline[0.6pt/1.6pt]
\addlinespace[0.25ex]
\multicolumn{21}{@{}l}{\textit{Reconstruction (ours)}} \\
\quad MotionMEM & 512+160 & 85.3 & 46.0 & 11.3 & 47.6 & 96.0 & 74.0 & 48.7 & 4.7 & 55.8 & 25.3 & 14.7 & 19.3 & 19.8 & 22.0 & 27.3 & 11.3 & 11.3 & 18.0 & 35.5 \\
\addlinespace[0.25ex]
\hdashline[0.6pt/1.6pt]
\addlinespace[0.25ex]
\multicolumn{21}{@{}l}{\textit{External}} \\
\quad NativeMEM & -- & 88.0 & 50.7 & 21.3 & 53.3 & 83.3 & 84.0 & 60.0 & 18.0 & 61.3 & 34.0 & 28.0 & 32.7 & 31.6 & 23.3 & 28.0 & 5.3 & 16.0 & \textbf{18.2} & 40.9 \\
\bottomrule
\end{tabular}
 }
\end{table*}

\begin{table*}[t]
  \centering
  \caption{\textbf{Success rates (\%, $\uparrow$) on the 16 RoboMME tasks~\cite{dai2026robomme}.} All methods inject the memory as context; $B$ is the history token budget. Each cell is the mean over three model seeds, and \cref{tab:main_std} adds the standard deviations. \textbf{Bold} marks the highest mean in each column.}
  \label{tab:main}
\setlength{\tabcolsep}{1.6pt}\renewcommand{\arraystretch}{1.05}\setlength{\aboverulesep}{0.2ex}\setlength{\belowrulesep}{0.3ex}
  \resizebox{\textwidth}{!}{\begin{tabular}{@{}llcccccccccccccccccccccc@{}}
\toprule
 &  &  & \multicolumn{5}{c}{\cellcolor{gray!15}Counting} & \multicolumn{5}{c}{\cellcolor{gray!15}Permanence} & \multicolumn{5}{c}{\cellcolor{gray!15}Reference} & \multicolumn{5}{c}{\cellcolor{gray!15}Imitation} &  \\
\cmidrule(lr){4-8}\cmidrule(lr){9-13}\cmidrule(lr){14-18}\cmidrule(lr){19-23}
Type & Method & $B$ & \makecell{Bin\\Fill} & \makecell{Pick\\Xtimes} & \makecell{Swing\\Xtimes} & \makecell{Stop\\Cube} & Avg & \makecell{Video\\Umsk} & \makecell{Button\\Umsk} & \makecell{Video\\UmskS} & \makecell{Button\\UmskS} & Avg & \makecell{Pick\\HighL} & \makecell{Video\\Repick} & \makecell{Video\\PlcBtn} & \makecell{Video\\PlcOrd} & Avg & \makecell{Move\\Cube} & \makecell{Insert\\Peg} & \makecell{Pattern\\Lock} & \makecell{Route\\Stick} & Avg & All \\
\midrule
Frames only & FrameSamp & 512 & \textbf{39.3} & \textbf{61.3} & 80.7 & 23.3 & 51.2 & 31.3 & 25.3 & 22.7 & 20.7 & 25.0 & 18.7 & 21.3 & 22.0 & 17.3 & 19.8 & 81.3 & 0.7 & 20.7 & 5.3 & 27.0 & 30.8 \\
 & FrameSamp & 1024 & 20.7 & 56.7 & 80.0 & 42.0 & 49.8 & 26.0 & 29.3 & 26.0 & 20.0 & 25.3 & \textbf{21.3} & 18.7 & 25.3 & 27.3 & 23.2 & 82.0 & \textbf{2.0} & 22.0 & 15.3 & 30.3 & 32.2 \\
\midrule
Ours & MotionMEM & 512+160 & 34.7 & 53.3 & \textbf{83.3} & \textbf{43.3} & \textbf{53.7} & \textbf{33.3} & \textbf{30.0} & \textbf{37.3} & \textbf{27.3} & \textbf{32.0} & 20.0 & \textbf{22.7} & \textbf{36.0} & \textbf{28.0} & \textbf{26.7} & \textbf{82.7} & \textbf{2.0} & \textbf{32.7} & \textbf{30.7} & \textbf{37.0} & \textbf{37.3} \\
\bottomrule
\end{tabular}
 }
\end{table*}

\noindent\textbf{Motion-centric tasks.}
We derive four tasks from RoboMME~\cite{dai2026robomme}: BinFill, RouteStick, VideoRepick, and VideoUnmaskSwap. In each, what must be remembered is a process shown over time, a route, a sequence of placements, or a swap, and no single frame shows it. Three of them gain an extra-hard tier, and BinFill gains a demonstration (\cref{sec:appendix_tasks}).  We generate 400 training demonstrations per task and evaluate on 700 held-out episodes in 14 task-and-difficulty groups, with seeds disjoint from training. These data are our own, and the results are not comparable with the RoboMME leaderboard. The motion encoder is pretrained on the same training demonstrations as the policies and sees no test episode. Policies are evaluated at 50k training steps with three evaluation seeds.

\noindent\textbf{General memory benchmark.}
For transfer we use the official RoboMME benchmark~\cite{dai2026robomme}: 16 memory-dependent manipulation tasks in four suites. We use the official split of 100 training demonstrations and 50 test episodes per task.  Every policy is fine-tuned for 60k steps with batch size 128, and we report the mean of three model seeds at the last checkpoint.

\noindent\textbf{Memory configuration.}
On the motion-centric tasks every frame contributes an $8\times8$ grid of 64 tokens, so the frame memories of 512, 1024, 2048, and 4096 tokens hold 8, 16, 32, and 64 evenly sampled frames. The budget sweep thus raises temporal density at a fixed spatial resolution. On RoboMME, the 512-token frame memory holds 32 frames of 16 tokens each.

\noindent\textbf{Baselines.}
We train every row of \cref{tab:longcontext} and \cref{tab:main}; within a table, rows share data, recipe, and schedule and differ only in the memory, except the external baseline. FrameSamp is the frame-sampling memory of RoboMME~\cite{dai2026robomme}, which MotionMEM keeps at 512 tokens. Wan-$k$ feeds frozen Wan latents to the policy as $k$ tokens per window, through a projection trained with the policy. Wan-1 uses the windows of MotionMEM and mean-pools the latents of each into one token. Wan-9 keeps all nine latents of each of at most 80 non-overlapping windows (720 tokens). E2E-Enc is our motion encoder on the same Wan latents, initialized randomly and trained only by the action loss. As an external baseline we train NativeMEM~\cite{wang2026nativemem}, one learned token per past frame, on the same data with its own recipe.  On the motion-centric tasks the main comparison uses both modulation and context injection; the budget sweep, the Wan-$k$ rows, and E2E-Enc use modulation only.

\section{Main Results}
\label{sec:rq1}

\subsection{Motion-Centric Tasks}
\label{sec:rq1_long}

\noindent\textbf{More frames saturate.}
Doubling the frame memory from 512 to 1024 tokens raises success from 31.7\% to 40.4\% (\cref{tab:longcontext}, \cref{fig:teaser}b). Two further doublings add four points and then almost nothing; the gain comes mostly from RouteStick, where denser snapshots trace the route.

\noindent\textbf{A few motion tokens outperform eight times the frames.}
MotionMEM keeps the 512 frame tokens and adds at most 160 motion tokens. It reaches 54.3\%, which is 9.7 points above the 4096-token frame memory with about a sixth of the tokens. At 4096 tokens the frame memory holds 64 frames, which on RouteStick and VideoRepick lie no more than about 16 steps apart, as dense as the motion tokens; the gap there is not one of temporal density. It is the highest on BinFill, RouteStick, and VideoRepick, with the widest margins where more frames helped little. VideoUnmaskSwap is the exception: under modulation every memory stays below 20\% there.

\noindent\textbf{Frozen video latents are not enough.}
Wan-1, at the same budget as MotionMEM, reaches 41.2\%, about 13 points below MotionMEM. On BinFill and RouteStick it is no better than the 2048-token frame memory: it behaves like more frames. Wan-9 keeps the latents unpooled, which raises the budget to 1232 tokens and lowers success to 37.6\%. Keeping the latents unpooled thus does not close the gap.

\noindent\textbf{Where the gain over end-to-end memory comes from.}
NativeMEM, the external baseline, learns its memory tokens end to end from sampled frames and reaches 40.9\%. E2E-Enc is also trained end to end but reads continuous video, and reaches 48.4\%. MotionMEM reads the same video and learns its token in a separate stage, by anchor-conditioned reconstruction, and reaches 54.3\%. Both steps help, on different tasks: E2E-Enc is about two points below MotionMEM on RouteStick but trails by 8 on BinFill and 16 on VideoRepick, where reconstruction makes the token describe every window. Wan-1 and E2E-Enc fill the same 160 slots, so extra positions do not explain the gain.

\noindent\textbf{Injection.}
Injecting the memory as context instead of by modulation lowers every row but keeps the overall gain: MotionMEM is about 14 points above the 512-token frame memory, the gap is again largest on RouteStick, and doubling the frames does not help. On VideoRepick alone MotionMEM falls below the frame memory (19.8\% vs.\ 27.1\%). The external baseline NativeMEM (40.9\%) stays above MotionMEM under context injection; it builds on a memory tokenizer pretrained on about 4,000 simulated and real-world trajectories~\cite{wang2026nativemem}.

\subsection{Transfer to General Memory Tasks}
\label{sec:rq1_transfer}

Doubling the frame memory from 512 to 1024 tokens changes the RoboMME average by 1.4 points (\cref{tab:main}). Adding at most 160 motion tokens to the 512-token memory instead reaches 37.3\%, vs.\ 32.2\% for 1024 frame tokens. Against the 1024-token memory it is higher in all four suites and on 13 of the 16 tasks. The tasks that gain at least ten points over both frame memories are decided by something that moves: the route of the stick on RouteStick, the traced pattern on PatternLock, the swap on VideoUnmaskSwap, and the event shown in the video on VideoPlaceButton. Tasks whose cue is visible in a single frame gain little, and InsertPeg stays near zero for every memory.

 \section{Analysis of the Motion Tokens}
\label{sec:interp}

\begin{figure}[t]
  \centering
  \includegraphics[width=\textwidth]{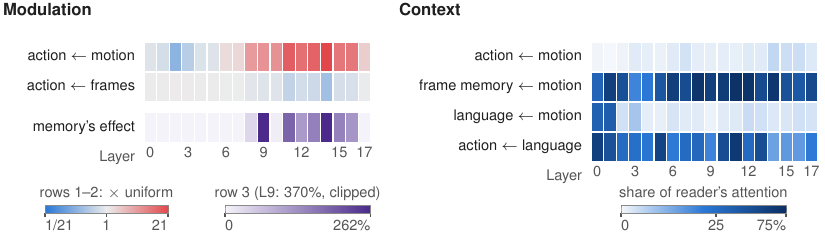}
  \caption{\textbf{Action tokens read motion tokens directly under modulation and through other tokens under context.} The 4-task MotionMEM policies on RouteStick, per layer. \emph{Left:} modulation, share of the memory attention on motion and frame tokens relative to uniform, and the relative change the memory makes to the modulation (clipped beyond the color range). \emph{Right:} context, share of each reading group's attention on each source group.}
  \label{fig:attn}
\end{figure}

\noindent\textbf{The action tokens read motion tokens directly.}
With modulation, the memory reaches the policy only through a memory attention in each action-expert layer. In layers 11 to 16 the action tokens put more than the uniform share of it on motion tokens and less on frame tokens, and the memory changes the modulation most in these layers and in layer 9 (\cref{fig:attn}, left). With context injection the memory tokens sit in the prefix, and the route differs (\cref{fig:attn}, right). In the context policy, the action tokens put about 1\% or less of their attention on motion tokens at every layer. The frame-memory tokens put 15 to 72\% of theirs on motion tokens, and the language tokens about a third in the first two layers; the action tokens in turn attend to the language tokens.

\begin{figure}[t]
  \centering
  \includegraphics[width=\textwidth]{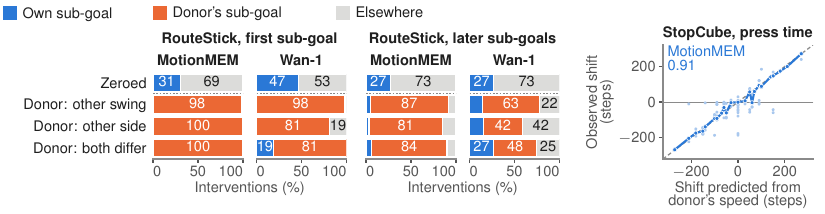}
  \caption{\textbf{Replacing motion tokens steers the policy.} \emph{Bars:} which sub-goal the robot executes on RouteStick when the motion tokens of one demonstrated swing are zeroed (above each dashed line) or replaced by a same-episode donor's (below), for MotionMEM and Wan-1 under modulation (4-task policies). \emph{Right:} StopCube press shift against the shift predicted from the donor's cube speed, for the 16-task context MotionMEM policy, with the fitted slope (line: median).}
  \label{fig:swap}
\end{figure}

\noindent\textbf{The policy conditions on what the token contains.}
On RouteStick, we replace the motion tokens of one demonstrated swing with tokens from the same episode, re-recorded with only that swing changed, at every sub-goal that the unmodified run reaches (\cref{fig:swap}). We do the same for the Wan-1 policy. At the first sub-goal both policies follow the donor. At later sub-goals, when only the swing differs, MotionMEM still takes the donor's swing in 87\% of runs, against 63\% for Wan-1. Zeroing the tokens instead changes the executed sub-goal in 67\% of interventions, against 8\% under a new sampling seed alone (\cref{sec:appendix_swap}). Replacing only the windows that lie entirely inside the swing has little effect on MotionMEM, so the effect comes from the windows that span its boundaries. On StopCube, the robot must press a button when a moving cube reaches the target for the $N$-th time~\cite{dai2026robomme}. We replace every motion token before the press with tokens recorded in the same scene at a different cube speed. The press shifts by 0.91 steps per step predicted from the donor's speed, although the frames show the true speed throughout.

\begin{figure}[t]
  \centering
  \includegraphics[width=\textwidth]{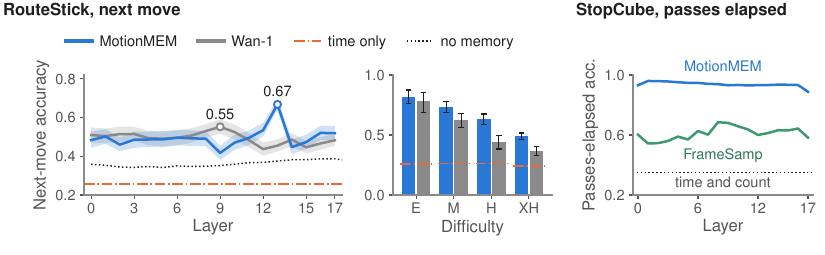}
  \caption{\textbf{Linear probes decode the next move on RouteStick and the passes elapsed on StopCube more accurately from the MotionMEM policy.} \emph{Left:} next-move accuracy from what the action tokens read from memory, per layer (4-task modulation policies; shaded: 95\% intervals; circles: selected layers). \emph{Middle:} the same at the selected layer, by difficulty. \emph{Right:} passes-elapsed accuracy from the language tokens (16-task context policies), on look-alike pairs one cube cycle apart; a pair counts only if both points are right. }
  \label{fig:belief}
\end{figure}

\noindent\textbf{The next move is decoded from what the policy reads from memory.}
We replay 200 new expert RouteStick episodes through the MotionMEM and Wan-1 policies and fit linear probes for the next move, one of four (\cref{fig:belief}). Time alone gives 26\%, and the prefix tokens, which cannot see the memory, 38\%. From what the action tokens read from memory it is decoded at 67\% for MotionMEM and 55\% for Wan-1, each at its best layer, with a gap of 11 to 19 points above the easy tier.  With context injection we probe StopCube, where the robot must know how many passes have elapsed. On pairs of points one cube cycle apart the images are near-identical, so a policy can tell them apart only from its history. Both points are decoded correctly in 89 to 96\% of pairs for the 16-task MotionMEM policy, against 54 to 68\% for FrameSamp. The comparisons are correlational.

\begin{figure}[t]
  \centering
  \includegraphics[width=4.6in]{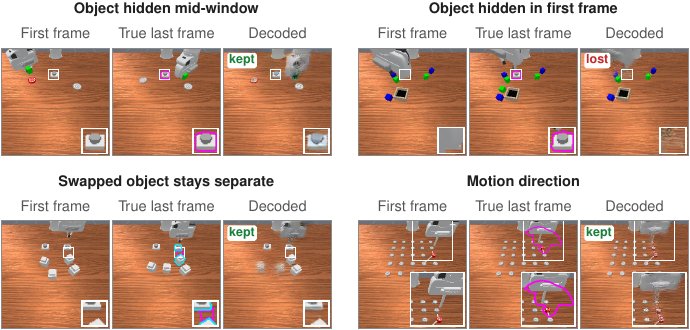}\caption{\textbf{Decoded motion tokens keep motion direction and object separation but lose an object hidden in the first frame.} For each event, left to right: first frame, true last frame with the judged region outlined, and last frame decoded from the motion token.}
  \label{fig:counting}
\end{figure}

\noindent\textbf{What the token keeps and what it loses.}
On held-out episodes, we decode each token to the last frame of its window with the pretraining decoder (\cref{fig:counting}). The decoded frame shows the true motion direction in all 579 windows, where the first frame alone does in none. An object that starts hidden and is revealed later is recovered in only 25\%. The decoder sees the first frame, so the token is never required to store what things look like, only how they move. The decoder is discarded before policy training, so these results describe the token, not what the policy reads from it.
 \section{Conclusion and Limitations}
\label{sec:conclusion}
We studied manipulation tasks in which what must be remembered is a process that no single frame shows. There, a sparse frame memory tells a VLA policy what the scene looked like, but not how it moved in between. MotionMEM keeps the frame memory and adds one motion token per window of video. Under modulation, on three of the four motion-centric tasks the token helps more than additional frames, and more than frozen or action-trained tokens; no memory helps on VideoUnmaskSwap. The gain also holds under context injection and on the RoboMME tasks. Replacing the tokens changes which sub-goal the policy executes, so the policy acts on what they record.

\noindent\textbf{Limitations.}
MotionMEM adds a video encoder to the control loop: its frozen video VAE makes inference about 4 times slower than FrameSamp (\cref{sec:appendix_longcontext}). The motion token complements the frame memory and cannot replace it, since it records change, not appearance. Our evidence is limited to one backbone, $\pi_{0.5}$, on simulated tasks derived from RoboMME. The encoder is trained on task data alone, and on RoboMME MotionMEM remains at least 21 points below methods that use an encoder pretrained on over 44k hours of video~\cite{sun2026ltm,gao2026dreamdojo}, a visual context of several thousand tokens~\cite{yin2026simplememvla}, or additional event annotations~\cite{liu2026ecomem}; these comparisons are uncontrolled.

\clearpage
\appendix

\section{Changes to the RoboMME Tasks}
\label{sec:appendix_tasks}
The four motion-centric tasks reuse the RoboMME environments~\cite{dai2026robomme} with the changes below. \cref{tab:task_tiers} lists the resulting 14 task-and-tier groups and their lengths in steps.

\noindent\textbf{RouteStick.}
We add an extra-hard tier that keeps the backtracking of the hard tier and raises the number of route segments from 4--7 to 8--10.

\noindent\textbf{VideoRepick.}
We add an extra-hard tier with the three cubes of the medium tier and 4--5 swaps instead of 2--3. We do not use the hard tier of RoboMME, so this task has three tiers.

\noindent\textbf{VideoUnmaskSwap.}
We add an extra-hard tier with the four containers of the hard tier and 4--5 swaps instead of 2--3.

\noindent\textbf{BinFill.}
BinFill has no demonstration in RoboMME. We add one: each episode is a successful execution played twice, first as the demonstration and then as the execution. The medium and hard tiers spawn 6--8 and 8--10 cubes instead of 8--10 and 10--12, and every target color receives at least one cube.

\begin{table}[h]
  \centering
  \caption{\textbf{The 14 task-and-tier groups.} $^{\dagger}$Tier that we added or whose setting we changed relative to RoboMME. Steps are over the training demonstrations; \emph{Demo.}\ is the part that must be remembered before execution starts. }
  \label{tab:task_tiers}
  \resizebox{\textwidth}{!}{\begin{tabular}{@{}lllrrr@{}}
\toprule
 & & & \multicolumn{3}{c}{Steps} \\
\cmidrule(l){4-6}
Task & Tier & Setting & Demo.\ mean & Episode mean & Episode max \\
\midrule
RouteStick & Easy & 2--3 segments & 125 & 250 & 300 \\
 & Medium & 4--5 segments & 225 & 450 & 500 \\
 & Hard & 4--7 segments, backtracking & 275 & 550 & 700 \\
 & Extra-hard$^{\dagger}$ & 8--10 segments, backtracking & 450 & 899 & 1{,}000 \\
\addlinespace[0.3ex]
VideoRepick & Easy & 3 cubes, 1--2 swaps & 296 & 688 & 922 \\
 & Medium & 3 cubes, 2--3 swaps & 347 & 746 & 968 \\
 & Extra-hard$^{\dagger}$ & 3 cubes, 4--5 swaps & 457 & 850 & 1{,}130 \\
\addlinespace[0.3ex]
VideoUnmaskSwap & Easy & 3 containers, 1--2 swaps & 141 & 327 & 517 \\
 & Medium & 4 containers, 1--2 swaps & 141 & 252 & 365 \\
 & Hard & 4 containers, 2--3 swaps & 192 & 455 & 552 \\
 & Extra-hard$^{\dagger}$ & 4 containers, 4--5 swaps & 291 & 562 & 1{,}312 \\
\addlinespace[0.3ex]
BinFill & Easy & 1 color, 4--6 cubes & 468 & 936 & 1{,}530 \\
 & Medium$^{\dagger}$ & 2 colors, 6--8 cubes & 642 & 1{,}284 & 1{,}762 \\
 & Hard$^{\dagger}$ & 3 colors, 8--10 cubes & 818 & 1{,}637 & 2{,}304 \\
\bottomrule
\end{tabular}
 }
\end{table}

\section{Evaluation Protocol}
\label{sec:appendix_longcontext}

\noindent\textbf{Memory configuration and protocol.}
The frame memory has 64 tokens per frame, with 8, 16, 32, and 64 frames for 512 to 4096 tokens. MotionMEM and Wan-1 add one token per 33-frame window, with windows every 16 frames and at most 160 windows. Wan-9 keeps the nine latents of each window as nine tokens, with windows every 32 frames and at most 80 windows. All policies are trained with batch size 128 and evaluated at 50k steps, with a limit of 2,000 steps per episode. The FrameSamp and MotionMEM policies were evaluated on A40 GPUs and the two Wan policies on A100 GPUs. \cref{tab:longcontext} gives the success rate of every task and tier.

\noindent\textbf{Inference cost.}
We time deployment on one A100. The policy predicts 20 actions and executes the first 16 before re-planning. Over 32 control steps, FrameSamp spends about 194 ms: two policy forwards of 67 ms each and about 60 ms to encode the new frames. For MotionMEM, the longer prefix and the motion encoder together add about 8\%, and the frozen video VAE that produces the window latents adds 0.59 s per window in bf16. With one new window in these 32 steps, they take about 4 times as long as for FrameSamp.

\section{Attention (\cref{fig:attn})}
\label{sec:appendix_attn}
Both halves of \cref{fig:attn} use 200 new expert RouteStick episodes that the policies did not see in training. We replay each episode through the policy and record its attention at points $t$ along the episode, 3,421 points in total. Every quantity below is computed at each point and layer $\ell$, averaged over the points of an episode, and then averaged over the 200 episodes with equal weight.

\noindent\textbf{Modulation.}
Each action-expert layer reads the memory through a memory attention. Let $\alpha^{\ell}_{t,k}$ be the weight that this attention puts on memory token $k$, averaged over heads, action tokens, and denoising steps, so that $\sum_{k=1}^{n_t}\alpha^{\ell}_{t,k}=1$ over the $n_t$ memory tokens. For the set $S$ of motion tokens, and likewise for the set of frame tokens, the first two rows report
\begin{equation*}
  r^{\ell}_{S}(t)=\frac{\sum_{k\in S}\alpha^{\ell}_{t,k}}{|S|/n_t},
\end{equation*}
the share of the attention on $S$ divided by the share that $S$ would receive under uniform attention, so $r^{\ell}_{S}>1$ means that the action tokens prefer $S$. The third row reports how much the memory changes the modulation,
\begin{equation*}
  \delta^{\ell}(t)=\frac{\lVert F^{\ell}(x,u)-F^{\ell}(x,0)\rVert}{\lVert F^{\ell}(x,0)\rVert},
\end{equation*}
where $F^{\ell}$ is the memory-conditioned normalization of layer $\ell$, $x$ is its input, and $u$ is the vector that the action tokens read from the memory.

\noindent\textbf{Context.}
The memory tokens sit in the prefix, and the policy reads them through the ordinary attention of the backbone. Let $A^{\ell}_{t}(i,k)$ be the attention weight from token $i$ to token $k$, averaged over heads. We group the tokens into frame-memory, motion, image, language, and action tokens, and for a reading group $Q$ and a source group $S$ we report
\begin{equation*}
  a^{\ell}_{Q\leftarrow S}(t)=\frac{1}{|Q|}\sum_{i\in Q}\sum_{k\in S}A^{\ell}_{t}(i,k),
\end{equation*}
the share of the attention of $Q$ that falls on $S$; for each $Q$ these shares sum to one over all sources. \cref{fig:attn} shows four pairs: action $\leftarrow$ motion, frame memory $\leftarrow$ motion, language $\leftarrow$ motion, and action $\leftarrow$ language.

\section{Token Replacement (\cref{fig:swap})}
\label{sec:appendix_swap}
The memory of the policy holds frame tokens and motion tokens, $\mathcal{M}=\{f_1,\dots,f_{B_{\mathrm{img}}}\}\cup\{m_1,\dots,m_J\}$, where $m_w$ is the motion token of window $w$ of the history. A replacement overwrites the motion tokens of a set of windows $\mathcal{R}$ and nothing else,
\begin{equation*}
  m'_w=\begin{cases} m^{\mathrm{don}}_w, & w\in\mathcal{R},\\ m_w, & w\notin\mathcal{R},\end{cases}
\end{equation*}
where $m^{\mathrm{don}}_w$ is the token of the same window in another recording, the donor, produced by the same encoder. The frame tokens, the current images, the instruction, and the environment are unchanged, so the policy sees the original episode everywhere except in the replaced motion tokens. Setting $m'_w=0$ for $w\in\mathcal{R}$ instead gives the zeroing intervention.

\noindent\textbf{RouteStick.}
The donor is the same episode re-recorded with only sub-goal $j$ of the demonstration changed: the swing direction flipped, the target moved to the other side, or both. $\mathcal{R}$ is the set of windows that overlap sub-goal $j$ in the demonstration. We then read which sub-goal the robot executes: its own, the donor's, or neither. On a pool of 60 new episodes this gives 923 interventions, zeroing included, for MotionMEM and 793 for Wan-1, both 4-task modulation policies. Zeroing changes the executed sub-goal, relative to the unmodified run, in 184 of the 275 zeroing interventions for MotionMEM (67\%); the bars of \cref{fig:swap} show instead which sub-goal is executed. As controls, a new sampling seed alone changes the executed sub-goal in 8\% of runs for MotionMEM and 19\% for Wan-1, and replacing only the windows that lie entirely inside sub-goal $j$ makes MotionMEM follow the donor in about 10\% of runs.

\noindent\textbf{StopCube.}
This test uses the 16-task context MotionMEM policy. The cube takes $M$ steps per pass, and the press is correct at its $N$-th visit to the target. The donor is the same scene recorded with the cube at another speed $M'$, and $\mathcal{R}$ is every window before the press, while the frames still show the true speed $M$. If the policy takes its timing from the motion tokens, the press should shift by $\Delta^{*}=(M'-M)(N-\tfrac12)$ steps. Over 72 episodes with $M\in\{60,80,120\}$ and $N\in\{2,\dots,5\}$, the observed shift regressed on $\Delta^{*}$ has slope 0.91 (95\% interval 0.84--0.97).

\section{Probes (\cref{fig:belief})}
\label{sec:appendix_belief}
A probe asks whether a label $y$ can be read linearly from a hidden vector $h\in\mathbb{R}^{d}$ of the policy. Each point $i$ of an episode gives one pair $(h_i,y_i)$. We split the episodes into five folds and predict every point with a probe fit on the other four. On these training folds we standardize $h$ and project it onto its first 64 principal components, $q=P^{\top}\big((h-\mu)/\sigma\big)$. For every class $c$ we then fit a ridge regressor to the indicator of the class,
\begin{equation*}
  (w_c,b_c)=\arg\min_{w,b}\;\sum_{i}\omega_i\big(w^{\top}q_i+b-\mathbf{1}[y_i=c]\big)^{2}+\lambda_c\lVert w\rVert^{2},\qquad \omega_i=\frac{1}{n_{e(i)}},
\end{equation*}
where $e(i)$ is the episode of point $i$ and $n_e$ its number of points, and $\lambda_c$ is chosen with four inner folds of the training episodes. The prediction is the class with the largest output, $\hat{y}_i=\arg\max_c\,(w_c^{\top}q_i+b_c)$. We report the accuracy on the held-out points with every episode weighted equally,
\begin{equation*}
  \mathrm{Acc}=\frac{1}{E}\sum_{e=1}^{E}\frac{1}{n_e}\sum_{i\in e}\mathbf{1}[\hat{y}_i=y_i].
\end{equation*}

\subsection{RouteStick: Next Move}
The label $y$ is the next move, one of four classes, at the 3,421 points of the 200 episodes of \cref{sec:appendix_attn}. We probe three vectors in every layer $\ell$: the prefix tokens (image and language), which cannot see the memory under modulation; the action tokens; and the vector $u$ that the action tokens read from the memory (\cref{sec:appendix_attn}), which is what \cref{fig:belief} reports. The time baseline replaces $h$ with the execution step alone. When a single layer is reported, it is selected inside the training folds; the selected layer is 13 for MotionMEM and 9 for Wan-1. Intervals are 95\% bootstrap intervals over episodes with 2,000 resamples.

\subsection{StopCube: Passes Elapsed}
This test uses the 16-task context policies of MotionMEM and FrameSamp. The label $y$ is the number of passes that the cube has completed, and $h$ is taken from the language tokens of each layer. Both policies replay the 100 official StopCube demonstrations, which are the training split, open-loop, and hidden states are recorded without executing actions; 98 demonstrations contain hover-phase points. We sample every eighth step of the hover phase and skip steps within two of a target crossing, which gives 2,344 points, identical for the two policies.

We score look-alike pairs: two points $(a,b)$ of one episode that are one cube cycle apart and whose images differ by at most 8 of 255 intensity levels at every pixel in both cameras. Their images are the same but their labels differ. 50 episodes contain 608 such pairs. With $\mathcal{P}_e$ the pairs of episode $e$, \cref{fig:belief} reports the fraction of pairs in which both points are decoded correctly, again with every episode weighted equally,
\begin{equation*}
  \mathrm{Acc}_{\mathrm{pair}}=\frac{1}{E'}\sum_{e=1}^{E'}\frac{1}{|\mathcal{P}_e|}\sum_{(a,b)\in\mathcal{P}_e}\mathbf{1}[\hat{y}_a=y_a]\,\mathbf{1}[\hat{y}_b=y_b],\qquad E'=50.
\end{equation*}

\section{Key-Event Decoding (\cref{fig:counting})}
\label{sec:appendix_keyevent}
The test asks what a motion token keeps of its 33-frame window. It uses validation episodes 94--99 of the encoder's training data. For a window with first frame $x_0$, true last frame $x$, and motion token $z$, the pretraining decoder gives the decoded last frame $\hat{x}=\mathrm{dec}(z,x_0)$.

\noindent\textbf{Score.}
To get a rate and not selected examples, each window is scored by a deterministic forced choice between the true frame $x$ and a counterfactual frame $\tilde{x}$ in which the event did not happen. Frames are compared inside the region $R$ of the event after a Gaussian blur $G_\sigma$ with $\sigma=1.5$,
\begin{equation*}
  d_R(u,v)=\frac{1}{|R|}\sum_{p\in R}\big\lVert (G_\sigma * u)(p)-(G_\sigma * v)(p)\big\rVert_2^{2},
\end{equation*}
the mean squared RGB distance over the pixels $p$ of $R$. A window passes if the decoded frame is strictly closer to the true frame than to every counterfactual,
\begin{equation*}
  d_R(\hat{x},x)<\min_{k}\,d_R(\hat{x},\tilde{x}_k).
\end{equation*}

\noindent\textbf{Events.}
Events are selected from simulator depth without looking at the decoded frames: back-projected depth gives the object regions, and the projected end-effector pose locates the arm. \emph{Occlusion:} $R$ is the object region at the last frame, and $\tilde{x}$ erases the object by inpainting. Windows are split by whether the object is visible in the first frame $x_0$. \emph{Swap:} $R$ is the gap between the swapped object and a static neighbor, and $\tilde{x}$ fills the gap with their mean color. \emph{Direction:} on PatternLock and RouteStick, the counterfactuals $\tilde{x}_k$ are the frame in which the end-effector stays and up to three real frames of the same episode that move by a similar distance (0.7 to 1.3 times) in a direction at least $45^{\circ}$ away.

\noindent\textbf{Results.}
\cref{tab:keyevent} gives the pass rates.  All thresholds were fixed before the decoded frames were scored. Each window shown in \cref{fig:counting} is the one closest to the median score for its event and task, not one selected for pass or fail.

\begin{table}[h]
  \centering
  \caption{\textbf{The motion token keeps motion direction, object separation, and objects seen in the first frame, but not objects hidden in the first frame.} Forced-choice pass rate (\%) on held-out episodes. \emph{VAE}: the true last frame through the VAE, an upper bound. \emph{First frame}: the first frame alone, a static picture. $^{\dagger}$Not a lower bound: the region is covered by the arm in the first frame.}
  \label{tab:keyevent}
  \footnotesize\setlength{\tabcolsep}{3.5pt}
  \begin{tabular}{@{}lrccc@{}}
  \toprule
  Event & $n$ & MotionMEM & VAE & First frame \\
  \midrule
  Object hidden mid-window & 79 & 93.7 & 100.0 & 100.0 \\
  Object hidden in first frame & 203 & 24.6 & 100.0 & --$^{\dagger}$ \\
  Swapped object stays separate & 54 & 100.0 & 100.0 & 92.6 \\
  Motion direction & 579 & 100.0 & 100.0 & 0.0 \\
  \bottomrule
  \end{tabular}
\end{table}

\section{Success Rates with Standard Deviations}
\makeatletter\setlength{\@fptop}{0pt}\makeatother \label{sec:appendix_std}
\begin{table}[p]
  \centering
  \caption{\textbf{Success rates (\%, $\uparrow$) on the four motion-centric tasks.} E, M, H, and XH are difficulty tiers; Avg pools a task and All pools the 700 episodes. $B$ is the history token budget. The upper and lower blocks inject the memory by modulation and as context. Cells are mean $\pm$ standard deviation over three evaluation seeds. \textbf{Bold}: highest per column.}
  \label{tab:longcontext_std}
  \fontsize{7.2pt}{7.6pt}\selectfont\renewcommand{\tiny}{\fontsize{4.6pt}{4.6pt}\selectfont\rule[-3.0pt]{0pt}{0pt}}\setlength{\tabcolsep}{3.0pt}\renewcommand{\arraystretch}{1.0}\renewcommand\cellset{\renewcommand\arraystretch{0.72}\setlength\extrarowheight{0pt}}\setlength{\aboverulesep}{0.2ex}\setlength{\belowrulesep}{0.3ex}
  {\fontsize{6.6pt}{6.6pt}\selectfont\setlength{\tabcolsep}{2.4pt}\renewcommand{\arraystretch}{1.12}\resizebox{\textwidth}{!}{\begin{tabular}{@{}llccccccccc@{}}
\toprule
 &  & \multicolumn{4}{c}{BinFill} & \multicolumn{5}{c}{RouteStick} \\
\cmidrule(lr){3-6}\cmidrule(lr){7-11}
Method & $B$ & E & M & H & Avg & E & M & H & XH & Avg \\
\midrule
\multicolumn{11}{@{}l}{\textbf{Modulation injection}} \\
\multicolumn{11}{@{}l}{\textit{Frames}} \\
\quad FrameSamp & 512 & 80.7\,$\pm$\,{5.0} & 57.3\,$\pm$\,{6.1} & 16.0\,$\pm$\,{3.5} & 51.3\,$\pm$\,{2.7} & 84.0\,$\pm$\,{3.5} & 44.7\,$\pm$\,{6.4} & 17.3\,$\pm$\,{2.3} & 0.7\,$\pm$\,{1.2} & 36.7\,$\pm$\,{1.4} \\
\quad FrameSamp & 1024 & 78.0\,$\pm$\,{2.0} & 52.7\,$\pm$\,{5.8} & 23.3\,$\pm$\,{2.3} & 51.3\,$\pm$\,{2.7} & 96.7\,$\pm$\,{1.2} & 88.0\,$\pm$\,{3.5} & 58.0\,$\pm$\,{2.0} & 5.3\,$\pm$\,{4.2} & 62.0\,$\pm$\,{0.5} \\
\quad FrameSamp & 2048 & 80.7\,$\pm$\,{3.1} & 68.0\,$\pm$\,{2.0} & 22.7\,$\pm$\,{3.1} & 57.1\,$\pm$\,{1.0} & \textbf{100.0}\,$\pm$\,{0.0} & 95.3\,$\pm$\,{2.3} & 70.0\,$\pm$\,{4.0} & 38.0\,$\pm$\,{3.5} & 75.8\,$\pm$\,{1.0} \\
\quad FrameSamp & 4096 & 86.7\,$\pm$\,{7.0} & 62.0\,$\pm$\,{6.9} & 30.0\,$\pm$\,{5.3} & 59.6\,$\pm$\,{6.0} & 98.7\,$\pm$\,{1.2} & \textbf{99.3}\,$\pm$\,{1.2} & 82.7\,$\pm$\,{2.3} & 28.0\,$\pm$\,{3.5} & 77.2\,$\pm$\,{1.4} \\
\addlinespace[0.25ex]
\hdashline[0.6pt/1.6pt]
\addlinespace[0.25ex]
\multicolumn{11}{@{}l}{\textit{Frozen latents}} \\
\quad Wan-1 & 512+160 & 85.3\,$\pm$\,{3.1} & 62.7\,$\pm$\,{3.1} & 20.7\,$\pm$\,{1.2} & 56.2\,$\pm$\,{1.0} & 94.0\,$\pm$\,{5.3} & 94.0\,$\pm$\,{2.0} & 51.3\,$\pm$\,{2.3} & 20.0\,$\pm$\,{6.0} & 64.8\,$\pm$\,{3.7} \\
\quad Wan-9 & 512+720 & 82.7\,$\pm$\,{3.1} & 52.7\,$\pm$\,{6.1} & 18.0\,$\pm$\,{0.0} & 51.1\,$\pm$\,{1.0} & 95.3\,$\pm$\,{1.2} & 87.3\,$\pm$\,{2.3} & 44.7\,$\pm$\,{4.2} & 3.3\,$\pm$\,{1.2} & 57.7\,$\pm$\,{0.8} \\
\addlinespace[0.25ex]
\hdashline[0.6pt/1.6pt]
\addlinespace[0.25ex]
\multicolumn{11}{@{}l}{\textit{End-to-end}} \\
\quad E2E-Enc & 512+160 & 88.7\,$\pm$\,{1.2} & 72.7\,$\pm$\,{6.1} & 32.0\,$\pm$\,{4.0} & 64.4\,$\pm$\,{2.3} & 98.7\,$\pm$\,{1.2} & 96.7\,$\pm$\,{2.3} & \textbf{85.3}\,$\pm$\,{1.2} & 46.0\,$\pm$\,{2.0} & 81.7\,$\pm$\,{0.3} \\
\addlinespace[0.25ex]
\hdashline[0.6pt/1.6pt]
\addlinespace[0.25ex]
\multicolumn{11}{@{}l}{\textit{Reconstruction (ours)}} \\
\quad MotionMEM & 512+160 & \textbf{92.7}\,$\pm$\,{4.2} & \textbf{86.7}\,$\pm$\,{2.3} & \textbf{38.7}\,$\pm$\,{7.0} & \textbf{72.7}\,$\pm$\,{1.2} & \textbf{100.0}\,$\pm$\,{0.0} & 94.7\,$\pm$\,{1.2} & 84.7\,$\pm$\,{3.1} & \textbf{56.0}\,$\pm$\,{7.2} & \textbf{83.8}\,$\pm$\,{2.0} \\
\midrule
\multicolumn{11}{@{}l}{\textbf{Context injection}} \\
\multicolumn{11}{@{}l}{\textit{Frames}} \\
\quad FrameSamp & 512 & 76.0\,$\pm$\,{6.0} & 25.3\,$\pm$\,{1.2} & 8.0\,$\pm$\,{5.3} & 36.4\,$\pm$\,{1.4} & 28.7\,$\pm$\,{4.2} & 16.7\,$\pm$\,{5.0} & 7.3\,$\pm$\,{2.3} & 0.0\,$\pm$\,{0.0} & 13.2\,$\pm$\,{2.1} \\
\quad FrameSamp & 1024 & 59.3\,$\pm$\,{5.0} & 13.3\,$\pm$\,{1.2} & 4.0\,$\pm$\,{2.0} & 25.6\,$\pm$\,{1.5} & 62.0\,$\pm$\,{6.9} & 42.0\,$\pm$\,{6.0} & 9.3\,$\pm$\,{2.3} & 0.0\,$\pm$\,{0.0} & 28.3\,$\pm$\,{1.4} \\
\addlinespace[0.25ex]
\hdashline[0.6pt/1.6pt]
\addlinespace[0.25ex]
\multicolumn{11}{@{}l}{\textit{Reconstruction (ours)}} \\
\quad MotionMEM & 512+160 & 85.3\,$\pm$\,{6.1} & 46.0\,$\pm$\,{4.0} & 11.3\,$\pm$\,{4.2} & 47.6\,$\pm$\,{4.0} & 96.0\,$\pm$\,{2.0} & 74.0\,$\pm$\,{2.0} & 48.7\,$\pm$\,{5.0} & 4.7\,$\pm$\,{2.3} & 55.8\,$\pm$\,{1.3} \\
\addlinespace[0.25ex]
\hdashline[0.6pt/1.6pt]
\addlinespace[0.25ex]
\multicolumn{11}{@{}l}{\textit{External}} \\
\quad NativeMEM & -- & 88.0\,$\pm$\,{5.3} & 50.7\,$\pm$\,{5.0} & 21.3\,$\pm$\,{3.1} & 53.3\,$\pm$\,{3.5} & 83.3\,$\pm$\,{3.1} & 84.0\,$\pm$\,{2.0} & 60.0\,$\pm$\,{4.0} & 18.0\,$\pm$\,{2.0} & 61.3\,$\pm$\,{1.0} \\
\bottomrule
\end{tabular}}
\par\vspace{6pt}
\resizebox{\textwidth}{!}{\begin{tabular}{@{}llcccccccccc@{}}
\toprule
 &  & \multicolumn{4}{c}{VideoRepick} & \multicolumn{5}{c}{VideoUnmaskSwap} &  \\
\cmidrule(lr){3-6}\cmidrule(lr){7-11}
Method & $B$ & E & M & XH & Avg & E & M & H & XH & Avg & All \\
\midrule
\multicolumn{12}{@{}l}{\textbf{Modulation injection}} \\
\multicolumn{12}{@{}l}{\textit{Frames}} \\
\quad FrameSamp & 512 & 24.0\,$\pm$\,{2.0} & 28.0\,$\pm$\,{4.0} & 22.7\,$\pm$\,{2.3} & 24.9\,$\pm$\,{1.4} & 18.0\,$\pm$\,{3.5} & 26.0\,$\pm$\,{3.5} & 6.7\,$\pm$\,{3.1} & \textbf{17.3}\,$\pm$\,{1.2} & 17.0\,$\pm$\,{2.2} & 31.7\,$\pm$\,{1.4} \\
\quad FrameSamp & 1024 & 27.3\,$\pm$\,{1.2} & 32.7\,$\pm$\,{3.1} & 31.3\,$\pm$\,{3.1} & 30.4\,$\pm$\,{0.8} & 23.3\,$\pm$\,{3.1} & 29.3\,$\pm$\,{5.0} & 7.3\,$\pm$\,{1.2} & 12.7\,$\pm$\,{3.1} & \textbf{18.2}\,$\pm$\,{1.0} & 40.4\,$\pm$\,{1.0} \\
\quad FrameSamp & 2048 & 30.0\,$\pm$\,{3.5} & 30.7\,$\pm$\,{4.2} & 24.0\,$\pm$\,{2.0} & 28.2\,$\pm$\,{2.3} & 16.0\,$\pm$\,{3.5} & 24.7\,$\pm$\,{2.3} & 6.7\,$\pm$\,{3.1} & 14.0\,$\pm$\,{2.0} & 15.3\,$\pm$\,{1.3} & 44.3\,$\pm$\,{1.0} \\
\quad FrameSamp & 4096 & 19.3\,$\pm$\,{3.1} & 26.7\,$\pm$\,{1.2} & 31.3\,$\pm$\,{1.2} & 25.8\,$\pm$\,{0.4} & 24.0\,$\pm$\,{2.0} & 22.7\,$\pm$\,{2.3} & 4.0\,$\pm$\,{2.0} & 9.3\,$\pm$\,{1.2} & 15.0\,$\pm$\,{0.5} & 44.6\,$\pm$\,{1.6} \\
\addlinespace[0.25ex]
\hdashline[0.6pt/1.6pt]
\addlinespace[0.25ex]
\multicolumn{12}{@{}l}{\textit{Frozen latents}} \\
\quad Wan-1 & 512+160 & 22.7\,$\pm$\,{1.2} & 30.0\,$\pm$\,{2.0} & 26.0\,$\pm$\,{5.3} & 26.2\,$\pm$\,{1.0} & 22.7\,$\pm$\,{5.0} & 26.0\,$\pm$\,{4.0} & 7.3\,$\pm$\,{3.1} & 14.0\,$\pm$\,{3.5} & 17.5\,$\pm$\,{1.5} & 41.2\,$\pm$\,{1.2} \\
\quad Wan-9 & 512+720 & 24.0\,$\pm$\,{0.0} & 24.7\,$\pm$\,{4.6} & 24.7\,$\pm$\,{2.3} & 24.4\,$\pm$\,{0.8} & 22.0\,$\pm$\,{3.5} & 24.7\,$\pm$\,{2.3} & 10.0\,$\pm$\,{2.0} & 12.0\,$\pm$\,{2.0} & 17.2\,$\pm$\,{1.6} & 37.6\,$\pm$\,{0.8} \\
\addlinespace[0.25ex]
\hdashline[0.6pt/1.6pt]
\addlinespace[0.25ex]
\multicolumn{12}{@{}l}{\textit{End-to-end}} \\
\quad E2E-Enc & 512+160 & 30.0\,$\pm$\,{2.0} & 16.7\,$\pm$\,{1.2} & 40.0\,$\pm$\,{6.9} & 28.9\,$\pm$\,{2.7} & \textbf{32.0}\,$\pm$\,{2.0} & 25.3\,$\pm$\,{2.3} & 6.0\,$\pm$\,{0.0} & 8.0\,$\pm$\,{2.0} & 17.8\,$\pm$\,{0.8} & 48.4\,$\pm$\,{0.6} \\
\addlinespace[0.25ex]
\hdashline[0.6pt/1.6pt]
\addlinespace[0.25ex]
\multicolumn{12}{@{}l}{\textit{Reconstruction (ours)}} \\
\quad MotionMEM & 512+160 & \textbf{48.7}\,$\pm$\,{5.0} & \textbf{36.7}\,$\pm$\,{2.3} & \textbf{50.0}\,$\pm$\,{4.0} & \textbf{45.1}\,$\pm$\,{3.2} & 16.0\,$\pm$\,{2.0} & \textbf{31.3}\,$\pm$\,{3.1} & \textbf{14.7}\,$\pm$\,{3.1} & 9.3\,$\pm$\,{2.3} & 17.8\,$\pm$\,{0.8} & \textbf{54.3}\,$\pm$\,{0.9} \\
\midrule
\multicolumn{12}{@{}l}{\textbf{Context injection}} \\
\multicolumn{12}{@{}l}{\textit{Frames}} \\
\quad FrameSamp & 512 & 33.3\,$\pm$\,{5.0} & 24.0\,$\pm$\,{5.3} & 24.0\,$\pm$\,{3.5} & 27.1\,$\pm$\,{2.0} & 24.7\,$\pm$\,{4.6} & 20.0\,$\pm$\,{3.5} & 10.0\,$\pm$\,{3.5} & 10.0\,$\pm$\,{4.0} & 16.2\,$\pm$\,{2.5} & 22.0\,$\pm$\,{1.0} \\
\quad FrameSamp & 1024 & 22.0\,$\pm$\,{2.0} & 11.3\,$\pm$\,{3.1} & 16.0\,$\pm$\,{2.0} & 16.4\,$\pm$\,{0.4} & 20.0\,$\pm$\,{2.0} & 22.7\,$\pm$\,{3.1} & 7.3\,$\pm$\,{3.1} & \textbf{17.3}\,$\pm$\,{4.2} & 16.8\,$\pm$\,{2.6} & 21.9\,$\pm$\,{0.9} \\
\addlinespace[0.25ex]
\hdashline[0.6pt/1.6pt]
\addlinespace[0.25ex]
\multicolumn{12}{@{}l}{\textit{Reconstruction (ours)}} \\
\quad MotionMEM & 512+160 & 25.3\,$\pm$\,{1.2} & 14.7\,$\pm$\,{7.0} & 19.3\,$\pm$\,{2.3} & 19.8\,$\pm$\,{3.2} & 22.0\,$\pm$\,{3.5} & 27.3\,$\pm$\,{5.0} & 11.3\,$\pm$\,{3.1} & 11.3\,$\pm$\,{1.2} & 18.0\,$\pm$\,{2.8} & 35.5\,$\pm$\,{0.7} \\
\addlinespace[0.25ex]
\hdashline[0.6pt/1.6pt]
\addlinespace[0.25ex]
\multicolumn{12}{@{}l}{\textit{External}} \\
\quad NativeMEM & -- & 34.0\,$\pm$\,{3.5} & 28.0\,$\pm$\,{0.0} & 32.7\,$\pm$\,{2.3} & 31.6\,$\pm$\,{1.7} & 23.3\,$\pm$\,{1.2} & 28.0\,$\pm$\,{2.0} & 5.3\,$\pm$\,{1.2} & 16.0\,$\pm$\,{2.0} & \textbf{18.2}\,$\pm$\,{0.8} & 40.9\,$\pm$\,{1.0} \\
\bottomrule
\end{tabular}}
}
 \end{table}

\begin{table}[p]
  \centering
  \caption{\textbf{Success rates (\%, $\uparrow$) on the 16 RoboMME tasks~\cite{dai2026robomme}.} All methods inject the memory as context; $B$ is the history token budget. Each cell is the mean $\pm$ standard deviation over three model seeds. \textbf{Bold} marks the highest mean in each column.}
  \label{tab:main_std}
  \fontsize{6.8pt}{7.2pt}\selectfont\renewcommand{\tiny}{\fontsize{4.6pt}{4.6pt}\selectfont\rule[-3.0pt]{0pt}{0pt}}\setlength{\tabcolsep}{0.8pt}\renewcommand{\arraystretch}{1.0}\renewcommand\cellset{\renewcommand\arraystretch{0.72}\setlength\extrarowheight{0pt}}\setlength{\aboverulesep}{0.2ex}\setlength{\belowrulesep}{0.3ex}
  {\fontsize{5.9pt}{5.9pt}\selectfont\setlength{\tabcolsep}{2.4pt}\renewcommand{\arraystretch}{1.12}\resizebox{\textwidth}{!}{\begin{tabular}{@{}lllcccccccccc@{}}
\toprule
 &  &  & \multicolumn{5}{c}{Counting} & \multicolumn{5}{c}{Permanence} \\
\cmidrule(lr){4-8}\cmidrule(lr){9-13}
Type & Method & $B$ & \makecell{Bin\\Fill} & \makecell{Pick\\Xtimes} & \makecell{Swing\\Xtimes} & \makecell{Stop\\Cube} & Avg & \makecell{Video\\Umsk} & \makecell{Button\\Umsk} & \makecell{Video\\UmskS} & \makecell{Button\\UmskS} & Avg \\
\midrule
Frames only & FrameSamp & 512 & \textbf{39.3}$\pm${3.1} & \textbf{61.3}$\pm${9.0} & 80.7$\pm${3.1} & 23.3$\pm${4.2} & 51.2$\pm${2.8} & 31.3$\pm${4.2} & 25.3$\pm${1.2} & 22.7$\pm${2.3} & 20.7$\pm${4.2} & 25.0$\pm${0.5} \\
 & FrameSamp & 1024 & 20.7$\pm${3.1} & 56.7$\pm${2.3} & 80.0$\pm${3.5} & 42.0$\pm${6.0} & 49.8$\pm${0.6} & 26.0$\pm${4.0} & 29.3$\pm${1.2} & 26.0$\pm${4.0} & 20.0$\pm${2.0} & 25.3$\pm${0.3} \\
\midrule
Ours & MotionMEM & 512+160 & 34.7$\pm${1.2} & 53.3$\pm${3.1} & \textbf{83.3}$\pm${2.3} & \textbf{43.3}$\pm${2.3} & \textbf{53.7}$\pm${0.3} & \textbf{33.3}$\pm${1.2} & \textbf{30.0}$\pm${6.0} & \textbf{37.3}$\pm${4.6} & \textbf{27.3}$\pm${3.1} & \textbf{32.0}$\pm${2.5} \\
\bottomrule
\end{tabular}}
\par\vspace{6pt}
\resizebox{\textwidth}{!}{\begin{tabular}{@{}lllccccccccccc@{}}
\toprule
 &  &  & \multicolumn{5}{c}{Reference} & \multicolumn{5}{c}{Imitation} &  \\
\cmidrule(lr){4-8}\cmidrule(lr){9-13}
Type & Method & $B$ & \makecell{Pick\\HighL} & \makecell{Video\\Repick} & \makecell{Video\\PlcBtn} & \makecell{Video\\PlcOrd} & Avg & \makecell{Move\\Cube} & \makecell{Insert\\Peg} & \makecell{Pattern\\Lock} & \makecell{Route\\Stick} & Avg & All \\
\midrule
Frames only & FrameSamp & 512 & 18.7$\pm${3.1} & 21.3$\pm${2.3} & 22.0$\pm${2.0} & 17.3$\pm${1.2} & 19.8$\pm${1.9} & 81.3$\pm${1.2} & 0.7$\pm${1.2} & 20.7$\pm${1.2} & 5.3$\pm${3.1} & 27.0$\pm${0.5} & 30.8$\pm${0.2} \\
 & FrameSamp & 1024 & \textbf{21.3}$\pm${3.1} & 18.7$\pm${2.3} & 25.3$\pm${1.2} & 27.3$\pm${3.1} & 23.2$\pm${1.3} & 82.0$\pm${3.5} & \textbf{2.0}$\pm${2.0} & 22.0$\pm${2.0} & 15.3$\pm${1.2} & 30.3$\pm${0.6} & 32.2$\pm${0.4} \\
\midrule
Ours & MotionMEM & 512+160 & 20.0$\pm${5.3} & \textbf{22.7}$\pm${3.1} & \textbf{36.0}$\pm${2.0} & \textbf{28.0}$\pm${5.3} & \textbf{26.7}$\pm${2.3} & \textbf{82.7}$\pm${3.1} & \textbf{2.0}$\pm${2.0} & \textbf{32.7}$\pm${1.2} & \textbf{30.7}$\pm${1.2} & \textbf{37.0}$\pm${0.9} & \textbf{37.3}$\pm${0.6} \\
\bottomrule
\end{tabular}}
}
 \end{table}
 \end{document}